# SpiRobs: Logarithmic Spiral-shaped Robots for Versatile Grasping Across Scales

**Zhanchi Wang,[1] Nikolaos M. Freris,[1,3,*] and Xi Wei[2,**]**

[1]School of Computer Science and Technology, University of Science and Technology of China, Hefei, Anhui, PRC, 230026.
[2]School of Chemistry and Materials Science, University of Science and Technology of China, Hefei, Anhui, PRC, 230026
[3]Lead contact
*Correspondence: nfr@ustc.edu.cn.
**Correspondence: wxi@ustc.edu.cn.

## SUMMARY

Realizing a soft manipulator with biologically comparable flexibility and versatility often requires careful selection of materials and actuation, as well as attentive design of its structure, perception, and control. Here, we report a new class of soft robots (SpiRobs) that morphologically replicates the logarithmic spiral pattern observed in natural appendages (e.g., octopus arms, elephant trunks, etc.). This allows for a common design principle across different scales and a speedy and inexpensive fabrication process. We further present a grasping strategy inspired by the octopus that can automatically adapt to a target object's size and shape. We illustrate the dexterity of SpiRobs and the ability to tightly grasp objects that vary in size by more than two orders of magnitude and up to 260 times self-weight. We demonstrate scalability via three additional variants: a miniaturized gripper (mm), a one-meter-long manipulator, and an array of SpiRobs that can tangle up various objects.

## INTRODUCTION

Certain animals have slender, flexible appendages, ranging from a few centimeters in length for seahorses and chameleons' prehensile tails[1,2] to more than a meter in length for octopus arms and elephant trunks[3,4]. These appendages are capable of remarkable complexity of movement and serve various important functions, such as prey capture, locomotion, manipulation, and defense. This has been a source of inspiration for the design and construction of soft continuum manipulators by exploiting soft materials or compliant mechanisms[5–7]. Although roboticists have successfully recreated flexible deformation in such robotic systems and have demonstrated their significant potential in handling fragile or irregularly shaped objects [8], safe human-robot interactive tasks[9–11], medical applications[12,13], etc., the biological examples still outperform engineer systems in terms of dexterity and agility. For example, elephant trunks can wrap a carrot with a diameter of 3 cm, while it can also grasp and stack 300 kg stumps over half a meter in diameter[14]. Octopus arms can reach out and catch a fish on a sub-second timescale[15].

The design of soft robots is one of the most important foundations for achieving bio-comparable movements and functions. Unlike rigid robots, which mainly use end effectors to manipulate objects, soft robots can use the entire body for object manipulation[16]. Therefore, the design of soft manipulators should consider not only the kinematics (i.e., how the central axis of the robot deforms)[17] but also the geometrics (i.e., the shape of the robot surface)[18] and, more importantly, the integration of the two. Due to the intrinsic compliance of soft materials, the deformation of these manipulators is hard to predict, such that the vast majority of existing prototypes are case-specific; the kinematics and geometrics are designed based on long simulations alongside trial-and-error[19–22].

We observed that despite the large diversity in terms of size (e.g., from millimeters to meters), anatomy (e.g., pure muscular or synergy of skeleton and muscles), and living environment (e.g., on land or in water), several animal appendages share a commonality in the form of a tightly packed shape that conforms to a logarithmic spiral (Figure 1A and S1)[23–25]. We believe this morphology is critical to their functions and thus may reveal clues for the synthetic design of the kinematics and geometrics of soft robots. Although the "spiral" shape (e.g., Archimedean spiral, logarithmic spiral, golden spiral, and some helical curves) has been purposefully used in the design of some soft manipulators[26–28] or has been observed as a result of deformation[29–33], a precise description of the kinematics and geometrics of the logarithmic spiral as a design principle is still lacking, in particular for how to incorporate the curling/uncurling of the spiral into the operation of soft manipulators.

The goal of this paper, therefore, is to go beyond specific animal examples and provide a precise description of the design, fabrication, and operation principles of a class of soft robots from a holistic geometric standpoint (the logarithmic spiral). This spiral morphology is particularly interesting because it is observed in a group of animal structures across various species and scales.

We present two principles to enable the recreation of logarithmic spiral-shaped curling/uncurling in an artificial system. First, we introduce a universal (across scales) design based on the discretizing and uncurling of the logarithmic spiral (Figure 1B and C). This allows us to effortlessly fabricate robots tailored to diverse scenarios (see Figure 1D). Second, we develop a grasping strategy inspired by the octopus, which enables the spiral-shaped body to reach and wrap around objects by controlling the curling/uncurling motion using cables. This strategy specifically capitalizes on the mechanical deformation upon contact to adapt to objects of variable shape and size (e.g., cup, pen, egg, strawberry, pineapple, and more), as well as to operate in confined spaces, with no need for precise feedback or complex planning & control. Besides, we demonstrate scalability via three applications: a miniaturized SpiRob for handling minuscule biological samples (e.g., an ant), a one-meter-long soft manipulator attached to a drone for performing dynamic grasping tasks, and an array of SpiRobs that can pick up various objects through entanglement.

## RESULTS

### Design of SpiRobs

The **logarithmic spiral** (also known as growth spiral) is represented in polar coordinates ($\rho$, $\theta$) by $\rho = ae^{b\theta}$, where $a$ and $b$ are scaling parameters. Our design is based on the discretization and uncurling of a logarithmic spiral via the following process (Figure 1B). We first construct what we call the ***central* spiral**, which is defined as the curve formed by the midpoints of the line segments connecting points on the spiral that differ in phase by $2\pi$: $\rho_c(\theta) = \frac{\rho(\theta)+\rho(\theta+2\pi)}{2}$. We then consider rays from the origin corresponding to fixed angle intervals ($\Delta\theta$): connecting the points of intersection with the original and central spiral forms a series of quadrilaterals. Finally, by mirroring these with respect to the central spiral, we form the discrete units of the robot's body (Figure 1B). Note that once the spiral parameters and discretization step ($\Delta\theta$) are given, the shape and size of the units are fully determined; in fact, adjacent units are in a fixed ratio (see EXPERIMENTAL PROCEDURES; 'Fabrication of SpiRobs'), which allows to simply design a single one and just scale it up/down to obtain a SpiRob of any scale. The uncurling of these units naturally results in a tapered body. This can be curled and wrapped back under the forces acting through a cable passing through the robot's body, with one end fixed to the tip through a knot and the other connected to the motor (Figure 1C). Through this process, the robot's kinematics (i.e., the deformability of the central axis) and geometrics (i.e., the shape and size) are fully determined and coupled with each other by the spiral equations (see EXPERIMENTAL PROCEDURES; 'Conceptual design of SpiRobs').

An elastic layer is inserted on the central axis that connects discrete units; this provides the restoring force for the robot to uncurl when the cable is relaxed. Extruding the 2D design pattern generates a 3D solid entity (Figure 2A, Step ①). Cutting it so that the length and width of each unit are approximately the same gives a multi-unit robot that is conical in 3D (Figure 2A, Step ②). Then, we punch holes of appropriate size for the cables to pass through in the final CAD drawing and fabricate the robot by 3D printing (Bambu Lab X1C, TPU@95A, see Figure S2A).

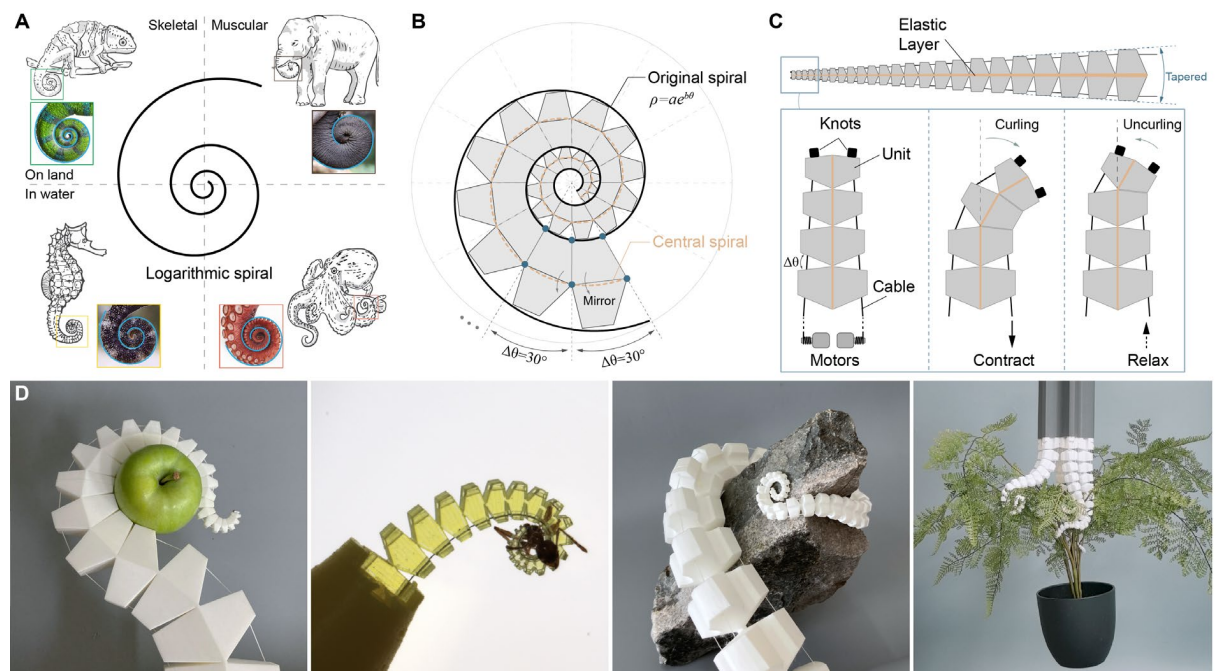

**Figure 1. Bioinspiration and design principle of spiral robots. (A)** Examples of animal parts that follow the logarithmic spiral. We provide the fitted spiral with the blue line. **(B)** Design of SpiRob: the logarithmic spiral ($\rho = ae^{b\theta}$) is discretized at a fixed step ($\Delta\theta$), and the resulting quadrilaterals are mirrored over the central spiral to obtain the robot units. **(C)** An elastic layer is added on the central axis. The contour of the robot body (dashed lines in the upper panel) forms a cone with a fixed taper angle. The cables are connected to motors and attached to the tipmost unit by knots. The cable contraction and relaxation are translated into the robot curling and uncurling. **(D)** Images of multiple SpiRobs: a two-cable robot holding an apple, a miniaturized robot grasping an ant, a three-cable robot wrapping around a rock, and an array of SpiRobs tangling up a plant. All robots were fabricated by 3D printing.

The units decrease in size from the base to the tip (at a fixed ratio; see Eq. (9) in MATERIALS AND METHODS). This is because the uniform discretization we apply is on the **angular** domain (i.e., $\Delta\theta$), which is key for the robot to curl into a logarithmic spiral. The radius of curvature (when curled) decreases from the base to the tip (it is affine to the length from the tip of the robot to that point; see Eq. (7), Table S1). This property is instrumental in SpiRobs wrapping and grasping a wide range of objects (especially small ones). It constitutes a critical difference with existing designs that mostly deform in a constant curvature manner[34].

We demonstrate the advantages of logarithmic spirals as a design principle through comparative experiments. We designed several prototypes (Figure S3A, three of which are shown in Figure 2B) with the same geometrics (i.e., size and shape of the contour of the robot body). Based on the SpiRob, we artificially reduce the gap between units by 10° without changing the length of each unit. This design reduces the local maximum bending of the robot, which we call SR-L (SpiRob of less deformation). In addition, we build a tapered robot following the constant curvature model (CC-T), whose base curvature is the same as that of SpiRob. We consider these robots to be two parts: one for grasping objects (like a hand, Figure 2B, gray part) and another for moving objects (like an arm, Figure 2B, blue part). For an object with a diameter $d$, we define the workspace as the set of all positions that can be grasped (the object is wrapped more than 1/2 circle). The maximum graspable workspace is obtained when the robot grasps the smallest object. The three prototypes can grasp the smallest object diameters are 13 mm, 32 mm, and 60 mm. Figure 2B shows that the SpiRob has the largest workspace; reducing the deformability reduces the workspace. The workspace for the CC-T robot is very limited, only about 1/5 of SpiRob.

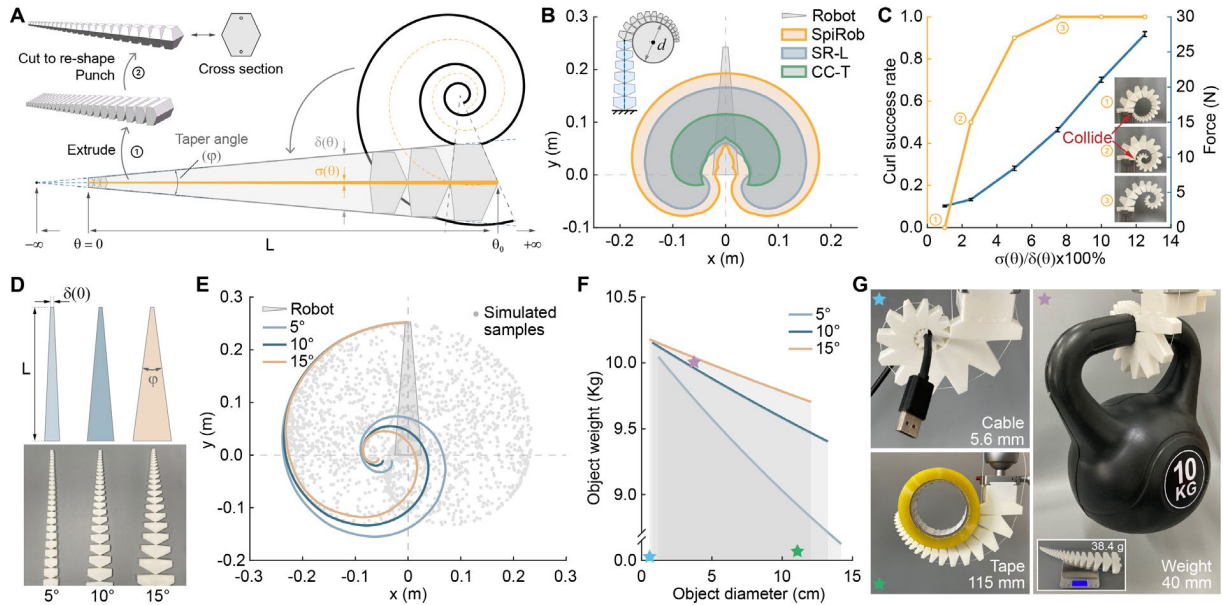

**Figure 2. Design parameters. (A)** A computer-aided design (CAD) model of a 25 cm long SpiRob driven by two cables: first, the 2D discretized pattern is extruded to obtain a 3D solid (Step ①), which is then cut into a conical shape, and two holes are reserved for the cables to pass through (Step ②). **(B)** Comparison of the workspaces of different designs when holding the smallest graspable object. SR-L: a SpiRob with less deformability. CC-T: a tapered constant-curvature robot. All robots have the same contour (size, taper angle, etc.), but the bodies are designed with different deformability. **(C)** Effect of elastic layer thickness on the curling motion of SpiRob. When the thickness is thin, the robot will collide with the base during curling (the cables are pulled slowly). Increasing the thickness can alleviate this problem, but it also means that a greater force is required to curl the robot into a spiral. **(D)** SpiRobs with different taper angles: 5°, 10°, and 15°; $L$ is the length of the robot and $\delta(0)$ is the width of its tip ($L = 25$ cm and $\delta(0) = 5.5$ mm in this case). **(E)** Workspace: the smaller the taper angle, the larger the workspace. The workspace envelopes are calculated using a kinematic model described in Supplemental Text S2. The gray dots illustrate the reachable points (10° robot) generated by randomly sampling cable forces in simulation based on a joint-link model (Text S4). **(F)** Theoretical predictions (see Text S2) of the object size and weight that can be grasped with a maximum actuation force of 100 N: a larger taper means a larger weight for a fixed diameter. **(G)** Images of a SpiRob (15°) grasping a 5.6 mm diameter cable, 115 mm diameter tape, and 10 kg weight. Stars in (F) and (G) are used to match the experiments to the weight-diameter plot.

Furthermore, we increased the gap between the units of SpiRob by 10° and designed SR-M (SpiRob of more deformation, Figure S3A). Although SR-M has a larger deformability than SpiRob, it can only curl into the same spiral state as SpiRob. In this case, it is not the local deformability that matters but the shape of the robot's body (the body is squeezed together). In the packing state, there are still gaps between the units, which reduces the stability. Generally, the gaps among the units in our design provide the space for the body to deform: when fully squeezed (i.e., two adjacent units touch), the robot can curl into a spiral shape (Video S1). Interestingly, there is an analogy with the elephant trunk, which features wrinkles and folds on its surface to support large deformations[43].

The thickness of the elastic layer is an important design choice for SpiRobs. The cables transmit actuation forces to the tip through knots and then affect the overall bending of the robot through the elastic layer. For quasi-static motion, the robot deforms synchronously along the body upon pulling the cable, and the thickness of the elastic layer affects the sequence of deformations at various locations. We conducted some pre-experiments (the cables are pulled slowly at a speed of 1 cm/s to curl the robot ten times from the rest configuration repeatedly) of different thicknesses (2.5% to 12.5% of the units' height, with an interval of 2.5%. A 1%-thickness prototype is used to simulate a situation where there is almost no elastic layer) to find a suitable value for designing the elastic layer: thinner elastic layers are likely to cause the robot fail to curl into a spiral due to self-collision during movement (Figure 2C, cases ① and ②) while thicker layers are less likely to curl (require higher force to curl into the spiral shape). Moreover, thinner layers are easier to buckle under antagonistic forces (Figure S3D and Video S2). In this paper, we chose a 5% thickness elastic layer as the design choice for the 2-cable spiral robots.

Another salient feature of our proposed design lies in the choice of the taper angle. We designed three SpiRobs with different taper angles (5°, 10°, and 15°) but the same length and tip diameter (Figure 2D). We found that the envelope of the workspace of a spiral robot also follows a spiral shape (Figure 2E, S4, and Text S2). Furthermore, the simulation in MuJoCo[35] validates that all interior points are reachable (Figure 2E, S5, and Text S3). In addition, we can analytically characterize the relationship between the size and weight of objects that can be grasped (Text S2). The following conclusions are drawn for fixed length and tip diameter. The smaller the taper angle, the larger the workspace (Figure 2E). At the same time, the larger the taper angle, the smaller the diameter of the smallest object that can be grasped and the larger the maximum load capacity; besides, for fixed diameter, the larger the weight (with the difference more pronounced for large-sized objects, see Figure 2F and Text S2). Taking the robot with a taper angle of 15° as an example, it can grasp objects whose diameters vary by two orders of magnitude (from 5.6 mm to 115 mm) and weigh roughly 260 times the weight of the robot (38.4 g self-weight and 10 Kg load capacity; Figure 2G, and Text S2 for the theoretical justification).

## Bioinspired grasping strategy

The second principle we leverage enables the active control of curling/uncurling to reach, wrap, grasp, and transport different objects. This movement pattern is reported in octopus[3,36]: it progressively uncurls and straightens curled arms to reach the target. After making contact with the object, the arm keeps uncurling on the surface of the object to align the suckers (Figure 3A and Video S2). In our system, we implement this by controlling the cable forces. Take the robot actuated by two cables as an example: when only the left or right cable is stretched, it can curl in the two corresponding directions to tightly pack into a spiral shape. By jointly controlling the forces exerted by the two cables, the robot reaches out, contacts the object, and uncurls the body along the object's surface to wrap around it and grasp it (Figure 3A, Video S3, and S6). We call the uncurling movement on the objects' surface "Climbing," similar to a plant (e.g., ivy) climbing on the wall. Similar grasping strategies are reported for grasping with multi-joint rigid robots[37]. Here, we show a much simpler implementation: antagonistically controlling the tension of two cables.

We have verified the effectiveness of this grasping and manipulation principle on a real robot (24 units for a total length of 45cm; see Figure 3B, S6, S7, and Text S5) and in simulations (where we use a model based on serial elastic joints). Unlike fingertip-based grasping paradigms[38], the proposed strategy utilizes all the robot surfaces to contact and wrap around the object. This is advantageous because a larger contact area means greater load capacity and grasping stability. We want to highlight that although our design does not include suckers like the octopus, uncurling the packed body on the surface of an object suffices for wrapping and grasping. Besides, during climbing, the robot's body does not slide relative to the contact surface, which is key for automatically adapting to objects of different roughness and shape. At the same time, this also enables the robot to easily navigate through confined spaces (the last two rows in Figure 3B, S8, and Video S7). We have further conducted a quantitative experiment to evaluate the graspable space of the proposed strategy for objects of different sizes. We divided the right half of the workspace into 5 cm × 5 cm grids, selected three objects of different diameters (25mm, 50mm, and 100mm), and carried out grasping tests at each grid point (Figure 3C). We found that the smaller the diameter, the closer the furthest point that can be grasped is to the boundary of the workspace. The same object may not be graspable if placed too far or close.

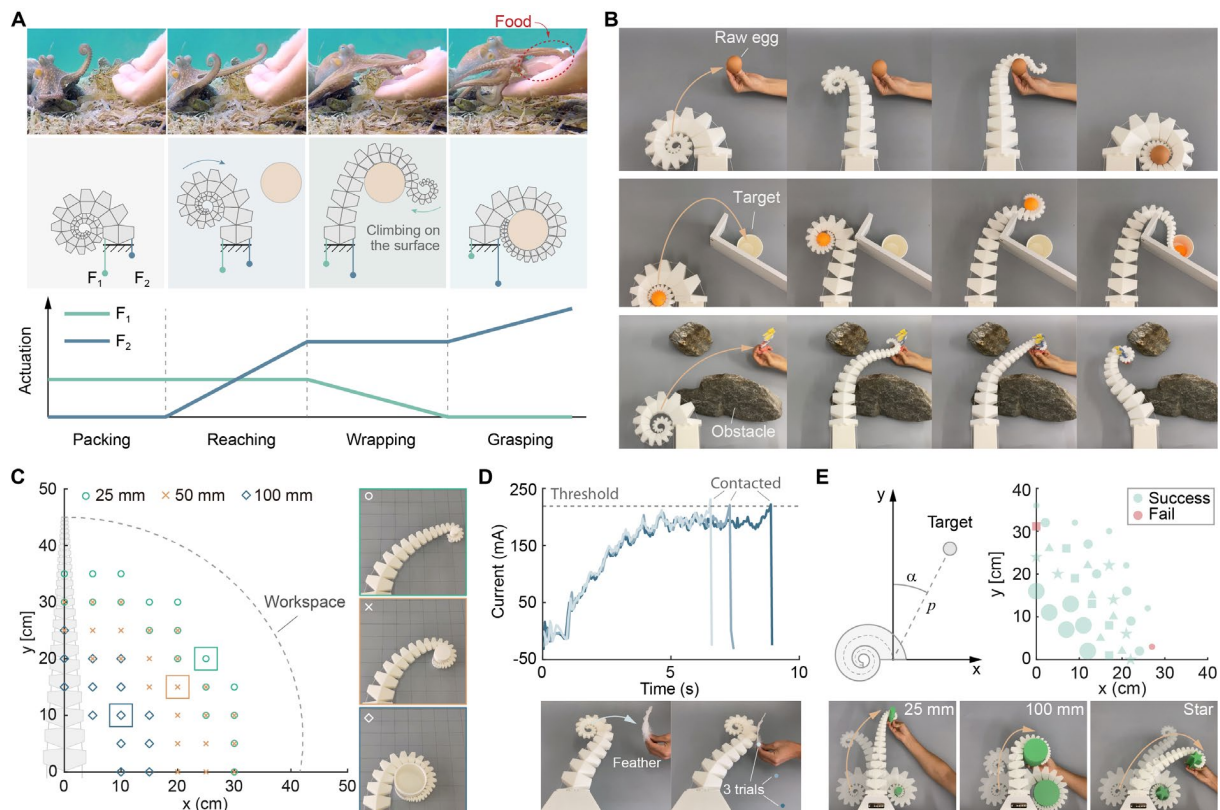

**Figure 3. Grasping. (A)** Snapshots of an octopus progressively uncurling its arm to reach, wrap, and grasp food (original video credits: *https://www.youtube.com/watch?v=GdCOYToDqfM*). Inspired by this, we implement a simple strategy to reproduce this motion pattern by controlling the force on two cables: starting from a state where the robot is packed into a spiral (*Packing*), increasing the force applied to the right cable ($F_2$) while keeping the left ($F_1$) unchanged translates to the robot starting to uncurl from the base and reaching out toward the object (*Reaching*). After being in contact with the object, by slowly reducing the force of the left cable while keeping the right unchanged, the robot uncurls (climbs) on the object's surface to wrap around it (*Wrapping*). After this, the force of the right cable is increased to achieve a firm grasp (*Grasping*). **(B)** SpiRobs grasping and moving various objects: grasping a raw egg, reaching through a crevice to grasp and retrieve a target object, and transporting a pingpong ball to the target behind the wall (more tests are provided in Figures S7 and S8). **(C)** Graspable space for different-sized objects with the proposed strategy. Colored points represent the locations where objects can be successfully grasped. The envelope of the workspace is plotted as a solid black line. **(D)** The robot senses the contact with a feather by detecting changes in the current. By repeating this across three trials (different positions), we determine that the same threshold can be used for contact detection across the workspace while maintaining high sensitivity. This detection mechanism can be used to initiate the reaching and wrapping stages of our strategy. **(E)** The robot automatically grasps objects of different geometries with minimal camera information of object positions ($p, \alpha$). Objects of different sizes are placed within the approximate area identified in (C).

To implement the grasping strategy automatically, we propose a simple yet effective approach based on motor current detection for switching between different actions. Starting from the *Packing* state, the robot uncurls and reaches out as the right cable is pulled, and the left cable maintains a constant force ($F_1^0$). The initial force on the left cable represents how tightly the robot is curled, affecting how easily the robot uncurls when the right cable is pulled. In our preliminary study[39], we modeled and analyzed the friction between the cable and the robot body to reproduce sequential deformations with a simple actuation input sequence, namely **piecewise-linear** actuation. Based on this working principle, we can control the robot to reach out in different directions by changing the initial force of the left cable ($F_1^0$) in the *Packing* state. For this purpose, we adopt a simple universal rule to define $F_1^0$ for objects at varying locations ($p, \alpha$) in polar coordinates on the 2D plane. In our implementation, this is given by a linear relationship $F_1^0 = -c_1 p + c_2 \alpha + c_0$, where $c_{0,1,2} > 0$ are the coefficients to be determined (determined experimentally as $c_0 = 14, c_1 = 13$, and $c_2 = 5$). The intuition behind this rule can be interpreted as follows: the farther the object is, the smaller the antagonistic force the robot is subjected to. Similarly, the larger the yaw angle of the object, the greater the antagonistic force.

After that, regarding the piecewise-linear actuation depicted in Figure 3A, the slope affects the speed of the robot's movement. We control the motor to work in different modes (speed mode or torque model) at different stages to obtain a stable movement. During the *Reaching* stage, the right cable is pulled at a constant speed. When there is contact with an object, the robot is subjected to an external force so that the motor current increases to maintain a constant speed of motion. Consequently, the robot can perceive contact by detecting such an increase in current. The threshold for contact detection was set through multiple experiments, where the maximum value of the current is recorded as the robot is running without contact. This simple mechanism is capable of detecting even the slightest contact with a feather (Figure 3D and Video S4). Then, the robot switches to the *Wrapping* stage, where the motor on the left gradually decreases the torque to zero. It thus slowly relaxes the left cable, letting the robot climb on the object's surface to wrap around it. When the torque of the left motor is zero, wrapping is completed. Then, the motor on the right pulls the cable at a constant speed to grasp the object and transfer it toward the robot's base. This method only requires the object's approximate location (e.g., as obtained by a camera, Kinect V2, Windows), circumventing the need for high-frequency feedback or precise motion planning, and can realize a high success rate of grasping objects of different shapes (six objects, each placed in six different positions, with a success rate of 94.4%. See Figure 3E and Video S5).

## Scalability
### 1. Small-scale SpiRobs
We scaled down the 2-cable robot in Figure 3 by 70 times to a length of about 1 cm across 29 units with a tip diameter of 0.14 mm (Figure 4A). We use a resin (ST1400, BMF) to fabricate the robot through stereolithography (SLA) 3D printing (MICROARCH S130, BMF). Two 20 $\mu m$-diameter cables pass through the robot's body and bond with resin on the tip-most unit (Figure 4C). They are controlled by two sliders of a pen-shaped handle (Figure 4B). We demonstrate that it can handle and manipulate small living organisms, i.e., an ant, without damage (Figure 4D and Video S14).

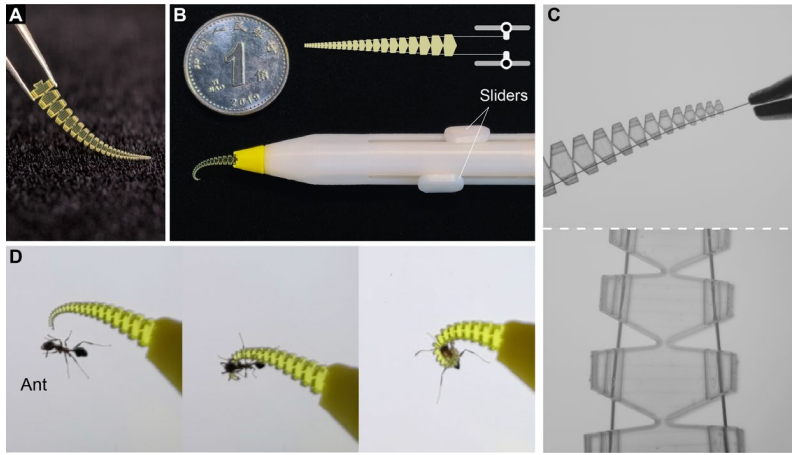

**Figure 4. SpiRob for non-invasive grasping of small living organisms. (A)** A 3D printed miniaturized (mm-scale) SpiRob. **(B)** System assembly. The two cables are connected to two sliders, and the robot can be curled in two directions by moving the sliders as in a multicolor pen. **(C)** Photographs of the cables and robot. **(D)** Delicate grasping of an ant with the miniaturized SpiRob.

## 2. 3-cable SpiRobs

We also designed and fabricated a one-meter-long robot (with a tip/base diameter of 5 mm/120 mm, 42 units) that is actuated by three cables evenly spaced on a circle corresponding to the cross-section (Figure 5A). We first obtain a series of parts by discretizing the spiral (as in Figure 1) and then generate a 3D entity by rotation (in place of mirroring used for the 2D case). An elastic axis (10% thickness) connects the discrete units. We further cut to reshape the cross-section (Figure 5A) for two purposes: a) to reduce the overall inertia and b) to increase the contact area when the robot grasps the object. The length of this robot exceeded the working area of the 3D printer, so we printed it in four parts and assembled them using dovetail connectors (Figure 5 B).

We manually operate the three cables with a strategy similar to the one for 2D grasping; the robot successfully unfolds the curled body on the object's surface to wrap, grasp, and manipulate it (Figure 5C, Video S15, and S16). For dynamic operation in the 3D space, higher flexibility can be achieved due to the intrinsic inertia and stiffness of the body and the effect of gravity. With a simple sawtooth actuation to one of the cables, the robot's body will curl up and then whip out to reach a specific position (Figure 5D, S13, Text S6, and Video S17). If the robot hits an object, the inertia of the motion will cause the tip of the robot to act like a whip (Figure 5D). Besides, an additional open-loop "curling" command (pull the other two cables) can be used to grasp an object dynamically (Figure 5E and Video S18). Since the whipping motion can sweep over the workspace (green shade in Figure 5E), the robot can make contact and grasp objects over a large area with these open-loop commands. In addition, the open-loop fashion in which we operate the robot enables it to move very fast, for example, grasp and lift a headset on the table within 1s.

Additionally, we attached this arm and its control terminal with three motors (Figure S2C and see MATERIALS AND METHODS section 'Control terminals') to a drone and conducted a grasping test with remote control (Figure 5B). The positions of the joystick are linearly mapped to the speed of the cables. In the experiment, the drone was controlled to hover near a plastic barrel. The SpiRob was then controlled to whip toward the handle and grasp it. Based on the onboard camera feedback, we command the robot to curl up to pick up the object (Figure 5F and Video S19). As expected from the SpiRob's adaptive qualities, the system can grasp the plastic barrel at a wide range, circumventing the need to hover precisely at a specific location.

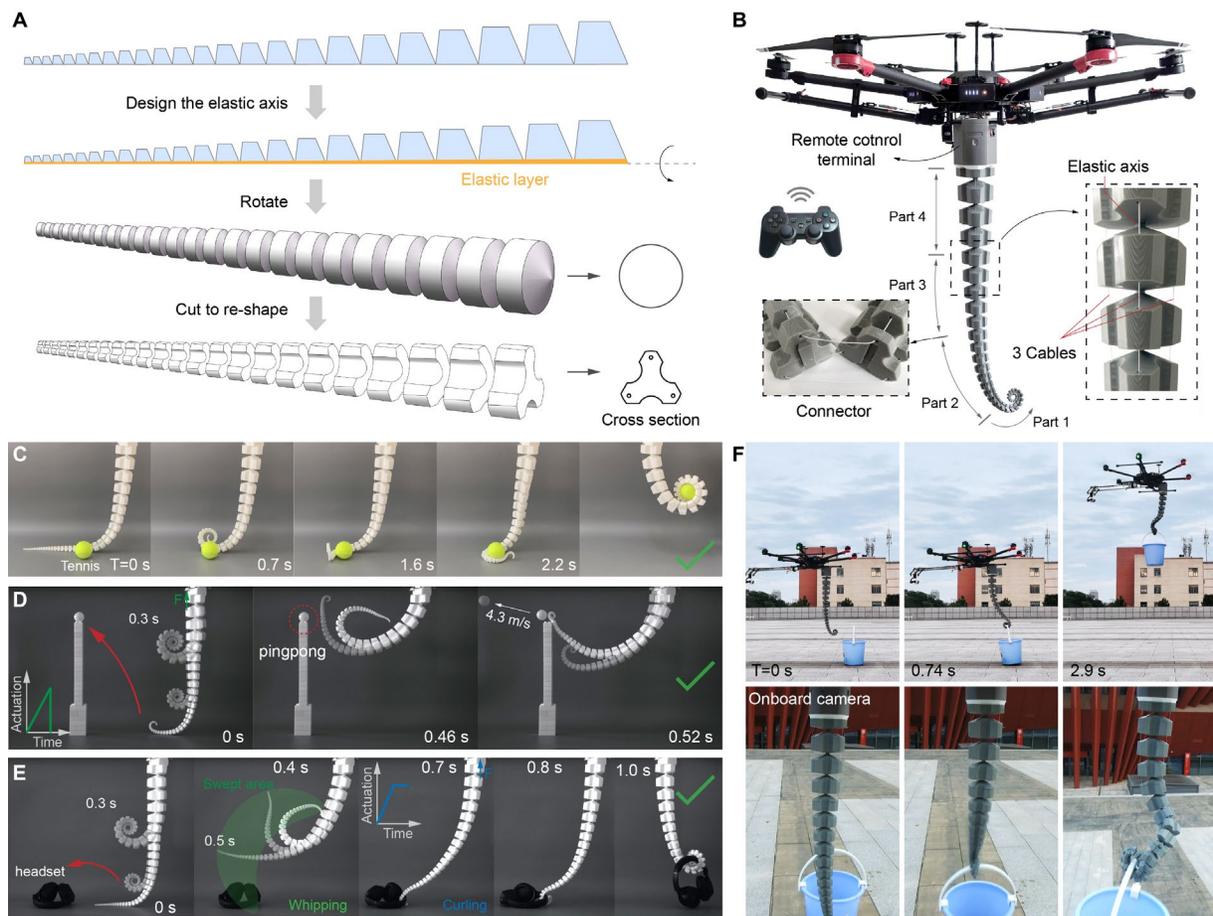

**Figure 5. Larger-scale 3-cable SpiRobs. (A)** Design of a 3D deformable SpiRob. We first rotate the parts around the central axis to form a cone-shaped body, which we then cut to obtain a cross-section that increases the contact area when wrapping around objects (while also reducing the total inertia). **(B)** Photographs of a 100 cm long 3-cable SpiRob and its installation to a drone. The robot comprises four 3D-printed parts assembled using dovetail connectors. A cylindric central axis serves as the elastic layer. **(C)** Image sequence of the robot picking up a ball from the table. **(D)** Whipping. The robot dynamically reaches a point on a sub-second timescale to pounce a pingpong ball with a simple sawtooth actuation. **(E)** The robot grasps and lifts a headset within 1 s. **(F)** Image sequence of the SpiRob grasping an object with a drone. Both the robot and drone are operated by remote control.

3. Multi-SpiRob array

We also built a gripper comprising multiple SpiRobs (each 3-cable, 25 cm in length with tip/base diameter of 5/30 mm), evenly arranged on a circumference (Figure 6A). We use this array to realize grasping via entanglement: this simple strategy applies to objects that are entirely different geometrically with minimal planning and no perception or feedback control (see Video S20). We attached the gripper to a rigid robotic arm and used sequences like those shown in Figure 6B for grasping. We first evaluated the effect of the number of arms on successfully grasping ten different objects (some are shown in Figure 6E, and others can be found in Table S2; five trials were conducted for each object). The results are shown in Figure 6C: when the number of SpiRobs in the array is six, the gripper can achieve a success rate of more than 90%. Increasing the number of arms further does not significantly improve the success rate (because a circle of larger diameter is needed to distribute the arms, so that grasping smaller objects, e.g., a tennis, is more likely to fail). We also measure the load capacity (defined as the force required to pull an object out of grasp) of the 6-SpiRob array using a force sensor. We artificially designed five tree-shaped targets with different numbers of branches (2, 4, 8, 8, and 16 branches, as shown in Figure 6D).

We found that the more complex the shape of the object, the more secure the grasp is, due to a higher level of intertwining. This is an interesting finding, given that objects like those we tested are often difficult for rigid robot grippers to handle. It is worth noting that when the shape complexity is too high (i.e., 16 branches in the experiment), there is not enough space for the gripper to fully entangle with the object, which reduces the load capacity. We conclude by pointing out that similar results, leveraging the robot's adaptability to simplify grasping with pneumatic filaments, were presented in [40,41]. Our design shows the ability to adapt to different objects through the active wrapping of objects, even for those with a size larger than the diameter of the gripper, such as the plastic pot (Figure 6E and Video S20).

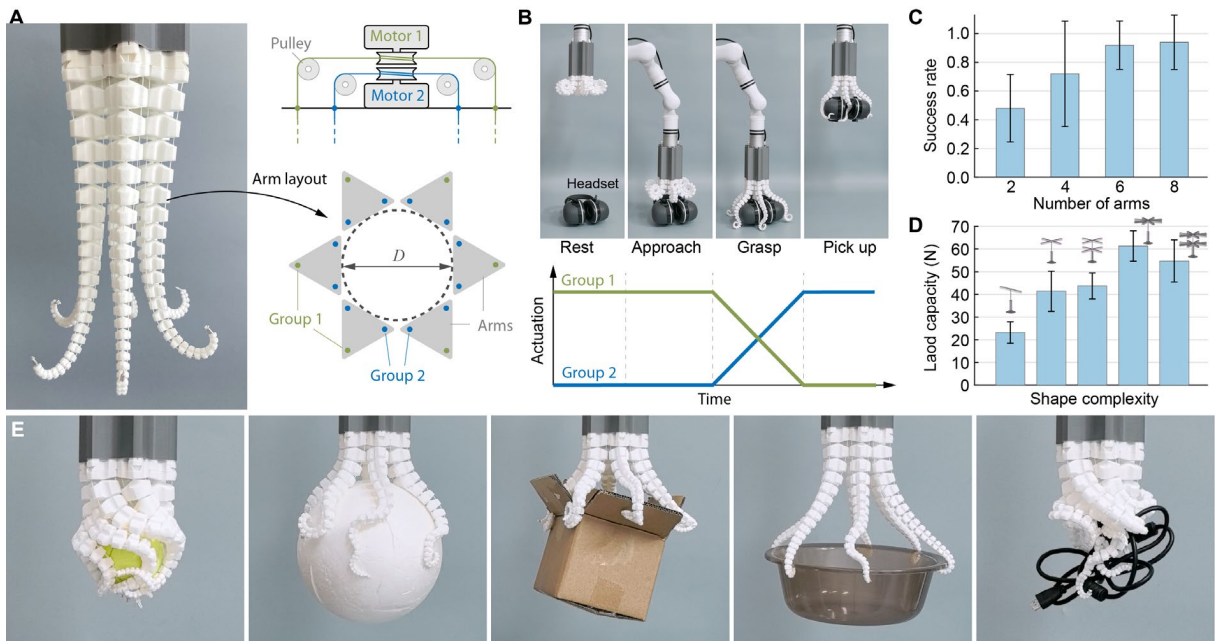

**Figure 6. An array of SpiRobs as a gripper. (A)** A multi-arm gripper consisting of 6 SpiRobs arms. The right panel shows the layout of the robots and how they are driven: the cables are divided into two groups: the outer (Group 1) and the inner (Group 2), each controlled by a motor. Pulling the outer cables causes all the robots to curl up, while pulling the inner cables causes the robots to uncurl. **(B)** The sequence of grasping with the array attached to a rigid robotic arm. The actuation patterns of the cables at different stages are plotted below. **(C)** The success rate of grasping experiments performed with different numbers of arms. Ten objects of different shapes and materials were used for testing, five of which are shown in (E). **(D)** The load capacity of the 6-arm gripper was measured in a test performed with various tree-shaped objects with increasing shape complexity (with an increased number of branches). **(E)** Photographs of an array of 6 SpiRobs actuated by a maximum force of 60 N to entangle around various objects.

## DISCUSSION

The design principle based on the logarithmic spiral constitutes the main novelty in this paper. Unlike most soft robots[6], where the hardware is designed first and the models are developed afterward, in our system, modeling (logarithmic spiral) comes first, and design/fabrication is a direct outcome of the model. This allows for designing SpiRobs for a large diversity of application scenarios (i.e., in terms of grasping size and load capacity, see also Table S1) instead of using a trial (building the robot) and error (testing) approach as well as resorting to extensive simulations.

The logarithmic spiral defines both the kinematics (i.e., deformation of the robot) and geometrics of the robot (i.e., taper angle, size, and shape). Unlike rigid robots whose kinematics can be fully described by a finite number of joint states, the deformation of soft manipulators is related to materials, structures, and actuation. This makes the design of soft manipulators lack a unified principle. Although we can use some tools (such as FEM) to optimize the design of soft manipulators, what is often lacking is not the means but the goal of optimization. The logarithmic spiral can potentially be used as a goal to examine whether the kinematics and geometrics are properly designed. In Figure S15, we show some representative soft manipulators and SpiRob for comparison. It can be seen that although many of them exhibit spiral or similar deformations, manipulators that achieve exact logarithmic spiral deformability in all directions have not yet been reported. In this paper, discretizing the spiral and using cables to actuate is a simple and easy-to-scale-up method. For SpiRob, both the kinematics and geometrics are related to the parameter $b$ of the spiral (Table S1), so they can be derived from each other (for a robot with a given taper angle and size; we can theoretically calculate how much deformability is needed at each location to be able to curl into a spiral).

The logarithmic spiral features large curvature changes along the body, thus resulting in a higher degree of flexibility close to the robot's tip. In contrast, most existing soft manipulators gain increased flexibility by combining multiple independently actuated segments; this, however, increases the complexity of modeling and control[42]. Based on a simple control strategy, we show that SpiRobs can support complex operations by using only two/three cables with minimal control effort. We also show how the robot's tip can be guided to a target position and orientation (Figure S10 and Video S9).

Instead of directly fabricating robots according to the spiral pattern, we uncurl it into a tapered shape. This facilitates the fabrication of the body with inexpensive 3D printing technology and simplifies the threading of cables. After uncurling, the paths of the cables are approximately straight lines. We also demonstrate an Origami fabrication of SpiRob with a sheet of paper (Figure S2D, E, and Text S1).

The bioinspired grasping strategy shows large adaptability and high grasping stability. It greatly increases the grasping space compared to direct bending (Figure S6). Even so, the grasping space is smaller than what the robot tip can reach since there is always a part of the body that is used to wrap around the object. To uncurl on the surface of an object, the SpiRob needs to make contact that exerts a force that may push it away. This was not an issue in the test cases with the 2-cable robot when objects were handed by a human. Still, we demonstrate that it is possible to grasp strawberries or pineapples that are not fixed initially (Figure S7). The 3-cable SpiRob further showcases the ability to grasp a ball placed on the table without bumping it away (Figure 5C). The multi-SpiRob array gripper greatly alleviates this problem, as multiple arms climb the object at the same time, thus enveloping the object.

Our design was inspired by natural creatures and operating strategies. To this end, we adopted a simplified abstraction from a geometric viewpoint (to morphologically capture the spiral characteristics). This allows for the replication of a rich kinematic and dynamic behavior with low complexity (2 or 3 cables). On this ground, interesting directions for future work include exploring to incorporate other grasping mechanisms (such as SpiRobs with suckers to mimic octopus arms[18] or an elephant trunk-like robot that can manipulate objects through wrapping, pinching, and sucking[4] and designs with independently-controllable arms (like the octopus).

## CONCLUSION

SpiRobs are bioinspired in both morphology and operation. A notable attribute is the scalability of the design principle (demonstrated with robots from mm to m scale). They are capable of complex movements and feature formidable adaptability in handling objects that vary in size and shape, with minimal sensing and actuation (2-3 cables). We further demonstrated three applications: a) a miniaturized robot that can handle fragile samples, b) teleoperation on a portable platform (drone), and c) a multi-arm gripper that functions via entanglement.

# EXPERIMENTAL PROCEDURES

## Resource availability
***Lead contact***: Further information and requests for resources and reagents should be directed to and will be fulfilled by the lead contact, Nikolaos M. Freris (nfr@ustc.edu.cn).
***Materials availability***: The materials can be produced following the procedures under Notes S3 and S5.
***Data and code availability***: All data from this study are available in the article and supplemental information.

## Conceptual design of SpiRobs
A SpiRob consists of three main components: the robot body, the cables, and the motors. The two ends of each cable are connected to the tipmost unit and the motor, and the cable's contract/relax actuation is respectively translated to the curling/uncurling motion of the robot. In the following, we describe the design process and determine the key parameters of the robot body.

We take a spiral robot with two cables as an example: this can wrap in two directions on a plane. Recall the expression of the logarithmic spiral (($\rho, \theta$) are polar coordinates):

$$\rho(\theta) = ae^{b\theta}, b = \cot\psi, \qquad (1)$$

where $a$ is a scaling factor, and $b$ is the cotangent of the constant polar tangential angle $\psi$ - defined as the angle between the tangent of a point on the spiral and the line connecting the point to the origin. Equation (1) can be re-written as $\rho = e^{b\left(\theta + \frac{\ln a}{b}\right)}$, so that $a$ can be interpreted as translation in the angle domain and $b$ as scaling. For design, we restrict attention to the range $\theta \geq 0$. The regime $\theta < 0$ corresponds to the extension from the tip to the point where the outer edges of the robot meet ($\theta \to -\infty$, as shown in Figure 2A). We define the 'central' spiral, which characterizes the central axis of the robot (Figure 1B), by:

$$\rho_c(\theta) = \frac{1}{2}\left(ae^{b(\theta+2\pi)} + ae^{b\theta}\right) = a_1 e^{b\theta}, a_1 = \frac{a}{2}(e^{2\pi b} + 1). \qquad (2)$$

The rays starting from the origin and at fixed angle intervals ($\Delta\theta = 30°$ in our case) intersect with points on the original spiral and the central spiral, connecting these to form quadrilaterals. This process builds one part of the robot, to which we attach an elastic layer that provides the restoring force that is crucial for the robot's flexibility. The other part is obtained by mirroring with respect to the central axis. In our design, the elastic layer's thickness is $5\% - 10\%$ of the unit width (5% for the 2-cable and 10% for the 3-cable robot in our implementation).

The taper angle ($\phi$) of the robot has the following relationship with the spiral parameters:

$$\phi = 2\arctan\left(\frac{\frac{1}{2}\delta(\theta)}{L(-\infty,\theta)}\right) = 2\arctan\left(\frac{b(e^{2\pi b}-1)}{\sqrt{b^2+1}\,(e^{2\pi b}+1)}\right). \qquad (3)$$

Where $\delta(\theta) = ae^{b(\theta+2\pi)} - ae^{b\theta}$ is the width of the robot at angle $\theta$ and

$$L(-\infty, \theta) = \int_{-\infty}^{\theta}\sqrt{\rho_c^2 + \dot\rho_c^2}\,d\theta = \frac{\rho_c(\theta)}{\cos\psi}, \qquad (4)$$

captures the length of the central spiral from the 'virtual' tip ($\theta \to -\infty$) to a given point (angle $\theta$) (see Figure 2A). The fact that $\phi$ in (3) is independent of $\theta$ shows that the spiral is tapered when expanded.

The length of the (central axis of the) robot has the following relationship with the spiral parameters:

$$L(0, \theta_0) = \int_0^{\theta_0}\sqrt{\rho_c^2 + \dot\rho_c^2}\,d\theta = \frac{\rho_c(\theta_0) - \rho_c(0)}{\cos\psi} = \frac{\sqrt{b^2+1}\,a_1}{b}\left(e^{b\theta_0} - 1\right). \qquad (5)$$

The lower bound of the integral here ($\theta = 0$) corresponds to the robot's tip, and the upper bound ($\theta_0$) to the root (Figure 2A). The curvature of the robot $\kappa(\theta)$ (when packed into a logarithmic spiral) can be calculated from the expression of the central spiral to be:

$$\kappa(\theta) = \frac{\sin\psi}{\rho_c(\theta)} = \frac{2}{a(e^{2\pi b}+1)\sqrt{b^2+1}}e^{-b\theta}. \qquad (6)$$

In particular, the curvature changes exponentially with $\theta$. We define the length $L(-\infty, \theta)$ in (4) as the local coordinate system $s$, and substitute it into (6) to eliminate $\theta$, then the radius of curvature $r(s)$ of any point on the spiral is given by:

$$r(s) = \frac{1}{\kappa(s)} = bs. \tag{7}$$

This shows that the radius of curvature of the central axis of SpiRob changes linearly. Thus, the Piecewise Constant Curvature (PCC) model, commonly used for modeling and controlling soft robots, is inappropriate for our case. Moreover, the deformation ratio ($\gamma$) -- defined as the ratio of the lengths of the robot's surfaces when the robot shifts from one packing state to its opposite packing state--can be calculated as:

$$\gamma = \frac{\int_{\theta_1+2\pi}^{\theta_2+2\pi} \sqrt{\rho^2+\dot{\rho}^2}\,d\theta}{\int_{\theta_1}^{\theta_2} \sqrt{\rho^2+\dot{\rho}^2}\,d\theta} = e^{2\pi b}, \tag{8}$$

For example, when $b = 0.22$ (i.e., taper angle $\phi = 15°$ in this case), $\gamma = 3.98$. The deformation ratio is independent of the start/end angle ($\theta_1, \theta_2$), which means that any part of the robot deforms at the same ratio. This is quite large for a continuous homogeneous body (e.g., from Silica gel) to undergo without deteriorating or even breaking from repeated stretching/shortening. Our multi-unit design (based on discretizing the spiral) is key for resolving this. This phenomenon also justifies the wrinkles and folds observed on the elephant trunk's surface, providing feasibility and sustainability for large deformations [43].

The design of a 3-cable robot that can curl in the 3D space follows the same principle as the planar 2-cable robot (Figure 5). Mirroring here is carried by rotation (which gives a cone). We further cut to reshape the cross-section to increase the contact area. A similar feature of a "square" cross-section that improves grasping has been observed in seahorse tails[1].

## Fabrication of SpiRobs

The robots are built using a desktop 3D printer (X1 Carbon, Bambu Lab) with 1.75 mm TPU (Thermoplastic polyurethanes) filament (eTPU-95A, Esun) (Figure S2A). Note that adjacent units of the robot are in a fixed ratio $\beta$, given by:

$$\beta = \frac{\delta(\theta+\Delta\theta)}{\delta(\theta)} = e^{b\Delta\theta}, \tag{9}$$

where $\Delta\theta$ is the discretization step. Thus, we only need to design one unit and then scale up/down according to the factor $\beta$ to obtain the adjacent ones. Second, longer robots can be fabricated in segments connected by a dovetail connector (Figure S3A). The 2-cable robot in Figure 3 consists of two segments, and the 3-cable robot in Figure 5 consists of four segments. We designed small holes on each unit for the cable to pass through. We use UHMWPE (ultra-high molecular weight polyethylene) cables for the robot's actuation. This cable type is wear-resistant and smooth, thus reducing friction as it moves through the robot's body. Furthermore, the cable is fastened to the tip unit with a fisherman knot to ensure that the actuation force is transmitted without slipping.

## Control terminals

We built three motor control terminals used to drive different robots. First, the terminal for 2-cable robot operation consists of two motors (GM6020, DJI), an embedded controller (Robomaster Development Board, type A, DJI), and a 24V battery. The motors are direct-drive brushless without a gearbox and can be controlled for torque, speed, and position. This also provides the basis for our current-based contact detection. Second, the terminal for 3-cable SpiRob (in the drone application) consists of three motors (M2006, DJI) and an embedded controller (Robomaster Development Board, type C, DJI). The terminal has a 2.4GHz wireless communication module for connecting to the joystick for remote control. The drone supplies the power to reduce the weight of the system. Finally, the terminal for multi-SpiRob grasping consists of two motors (M2006, DJI), a microcontroller (Robomaster Development Board, type C, DJI), and a power supply placed in the base of the rigid robotic arm.

### Data analysis

The motor current, speed, and position data are captured using STM32Cube Monitor. The cable length is converted from the recorded rotor position of the motors in MATLAB. For dynamic tasks (Figure 5), videos and screenshots are captured using a high-speed camera (ACS-3, NAC Image Technology).

## SUPPLEMENT INFORMATION

Notes S1 to S7 for multiple supplemental materials and methods.
Tables S1 and S2 for supplemental tables
Figures S1 to S15 for multiple supplemental figures
Videos S1 to S20 for multiple supplemental Videos

*Supplemental video titles:*
Video S1: 1-cable spiral robots
Video S2: Effect of design parameters
Video S3: Grasping strategy
Video S4: Contact detection
Video S5: Automatic grasping
Video S6: Grasping and manipulation of various objects
Video S7: Manipulation in confined space
Video S8: Remote control
Video S9: Controllability of the tip
Video S10: Dynamic motion of the 2-cable robot
Video S11: Grasping high-speed moving objects
Video S12: Impact-resistant grasping
Video S13: Elephant throw
Video S14: Miniature SpiRob
Video S15: 3-cable spiral robot
Video S16: Non-slippable grasping
Video S17: Whipping
Video S18: Whipping to grasp
Video S19: Application with a drone
Video S20: Multi-SpiRob gripper

## ACKNOWLEDGMENTS

## AUTHOR CONTRIBUTIONS

## DECLARATION OF INTERESTS

The authors declare that they have no competing financial interests. Z.W. and N.M.F. are listed as inventors on a Chinese patent application (CN114770585B) submitted by the University of Science and Technology of China that covers the fundamental principles and designs of SpiRobs.

## REFERENCES

1.  Porter, M.M., Adriaens, D., Hatton, R.L., Meyers, M.A., and McKittrick, J. (2015). Why the seahorse tail is square. Science *349*, aaa6683. https://doi.org/10.1126/science.aaa6683.

2.  Luger, A.M., Ollevier, A., De Kegel, B., Herrel, A., and Adriaens, D. (2020). Is variation in tail vertebral morphology linked to habitat use in chameleons? Journal of Morphology *281*, 229–239. https://doi.org/10.1002/jmor.21093.

3. Kier, W.M., and Stella, M.P. (2007). The arrangement and function of octopus arm musculature and connective tissue. Journal of Morphology *268*, 831–843. https://doi.org/10.1002/jmor.10548.

4. Dagenais, P., Hensman, S., Haechler, V., and Milinkovitch, M.C. (2021). Elephants evolved strategies reducing the biomechanical complexity of their trunk. Current Biology *31*, 4727-4737.e4. https://doi.org/10.1016/j.cub.2021.08.029.

5. Kim, S., Laschi, C., and Trimmer, B. (2013). Soft robotics: a bioinspired evolution in robotics. Trends in Biotechnology *31*, 287–294. https://doi.org/10.1016/j.tibtech.2013.03.002.

6. Rus, D., and Tolley, M.T. (2015). Design, fabrication and control of soft robots. Nature *521*, 467–475. https://doi.org/10.1038/nature14543.

7. Laschi, C., Mazzolai, B., and Cianchetti, M. (2016). Soft robotics: Technologies and systems pushing the boundaries of robot abilities. Sci. Robot. *1*, eaah3690. https://doi.org/10.1126/scirobotics.aah3690.

8. Ilievski, F., Mazzeo, A.D., Shepherd, R.F., Chen, X., and Whitesides, G.M. (2011). Soft Robotics for Chemists. Angewandte Chemie International Edition *50*, 1890–1895. https://doi.org/10.1002/anie.201006464.

9. Ansari, Y., Manti, M., Falotico, E., Mollard, Y., Cianchetti, M., and Laschi, C. (2017). Towards the development of a soft manipulator as an assistive robot for personal care of elderly people. International Journal of Advanced Robotic Systems *14*, 172988141668713. https://doi.org/10.1177/1729881416687132.

10. Jiang, H., Wang, Z., Jin, Y., Chen, X., Li, P., Gan, Y., Lin, S., and Chen, X. (2021). Hierarchical control of soft manipulators towards unstructured interactions. The International Journal of Robotics Research *40*, 411–434. https://doi.org/10.1177/0278364920979367.

11. Guan, Q., Stella, F., Della Santina, C., Leng, J., and Hughes, J. (2023). Trimmed helicoids: an architectured soft structure yielding soft robots with high precision, large workspace, and compliant interactions. npj Robot *1*, 4. https://doi.org/10.1038/s44182-023-00004-7.

12. Burgner-Kahrs, J., Rucker, D.C., and Choset, H. (2015). Continuum Robots for Medical Applications: A Survey. IEEE Transactions on Robotics *31*, 1261–1280. https://doi.org/10.1109/TRO.2015.2489500.

13. Russo, M., Sadati, S.M.H., Dong, X., Mohammad, A., Walker, I.D., Bergeles, C., Xu, K., and Axinte, D.A. (2023). Continuum Robots: An Overview. Advanced Intelligent Systems, 2200367. https://doi.org/10.1002/aisy.202200367.

14. Wilson, J.F., Mahajan, U., Wainwright, S.A., and Croner, L.J. (1991). A Continuum Model of Elephant Trunks. Journal of Biomechanical Engineering *113*, 79. https://doi.org/10.1115/1.2894088.

15. Gutfreund, Y., Flash, T., Yarom, Y., Fiorito, G., Segev, I., and Hochner, B. (1996). Organization of Octopus Arm Movements: A Model System for Studying the Control of Flexible Arms. J. Neurosci. *16*, 7297–7307. https://doi.org/10.1523/JNEUROSCI.16-22-07297.1996.

16. Walker, I.D., Dawson, D.M., Flash, T., Grasso, F.W., Hanlon, R.T., Hochner, B., Kier, W.M., Pagano, C.C., Rahn, C.D., and Zhang, Q.M. (2005). Continuum robot arms inspired by cephalopods. In, G. R. Gerhart, C. M. Shoemaker, and D. W. Gage, eds., p. 303. https://doi.org/10.1117/12.606201.

17. Calisti, M., Arienti, A., Giannaccini, M.E., Follador, M., Giorelli, M., Cianchetti, M., Mazzolai, B., Laschi, C., and Dario, P. (2010). Study and fabrication of bioinspired Octopus arm mockups tested on a multipurpose platform. In 2010 3rd IEEE RAS EMBS International Conference on Biomedical Robotics and Biomechatronics, pp. 461–466. https://doi.org/10.1109/BIOROB.2010.5625959.

18. Xie, Z., Domel, A.G., An, N., Green, C., Gong, Z., Wang, T., Knubben, E.M., Weaver, J.C., Bertoldi, K., and Wen, L. (2020). Octopus Arm-Inspired Tapered Soft Actuators with Suckers for Improved Grasping. Soft Robotics *7*, 639–648. https://doi.org/10.1089/soro.2019.0082.

19. Cianchetti, M., Arienti, A., Follador, M., Mazzolai, B., Dario, P., and Laschi, C. (2011). Design concept and validation of a robotic arm inspired by the octopus. Materials Science and Engineering: C *31*, 1230–1239. https://doi.org/10.1016/j.msec.2010.12.004.

20. Morales Bieze, T., Kruszewski, A., Carrez, B., and Duriez, C. (2020). Design, implementation, and control of a deformable manipulator robot based on a compliant spine. The International Journal of Robotics Research *39*, 1604–1619. https://doi.org/10.1177/0278364920910487.

21. Jiang, H., Liu, X., Chen, X., Wang, Z., Jin, Y., and Chen, X. (2016). Design and simulation analysis of a soft manipulator based on honeycomb pneumatic networks. In 2016 IEEE International Conference on Robotics and Biomimetics (ROBIO), pp. 350–356. https://doi.org/10.1109/ROBIO.2016.7866347.

22. Xie, Z., Yuan, F., Liu, J., Tian, L., Chen, B., Fu, Z., Mao, S., Jin, T., Wang, Y., He, X., et al. (2023). Octopus-inspired sensorized soft arm for environmental interaction. Sci. Robot. *8*, eadh7852. https://doi.org/10.1126/scirobotics.adh7852.

23. Thompson, D.W. (1992). On Growth and Form 1st ed. J. T. Bonner, ed. (Cambridge University Press) https://doi.org/10.1017/CBO9781107325852.

24. Porter, M.M., Novitskaya, E., Castro-Ceseña, A.B., Meyers, M.A., and McKittrick, J. (2013). Highly deformable bones: Unusual deformation mechanisms of seahorse armor. Acta Biomaterialia *9*, 6763–6770. https://doi.org/10.1016/j.actbio.2013.02.045.

25. Hammer, Ø. (2016). The Perfect Shape (Springer International Publishing) https://doi.org/10.1007/978-3-319-47373-4.

26. Zhang, Z., Wang, X., Meng, D., and Liang, B. (2021). Bioinspired Spiral Soft Pneumatic Actuator and Its Characterization. J Bionic Eng *18*, 1101–1116. https://doi.org/10.1007/s42235-021-00075-y.

27. Zournatzis, I., Kalaitzakis, S., and Polygerinos, P. (2023). SoftER: A Spiral Soft Robotic Ejector for Sorting Applications. IEEE Robot. Autom. Lett. *8*, 7098–7105. https://doi.org/10.1109/LRA.2023.3315206.

28. Taylor, I.H., Bawa, M., and Rodriguez, A. (2023). A Tactile-enabled Hybrid Rigid-Soft Continuum Manipulator for Forceful Enveloping Grasps via Scale Invariant Design. In 2023 IEEE International Conference on Robotics and Automation (ICRA), pp. 10331–10337. https://doi.org/10.1109/ICRA48891.2023.10161121.

29. Paek, J., Cho, I., and Kim, J. (2015). Microrobotic tentacles with spiral bending capability based on shape-engineered elastomeric microtubes. Sci Rep *5*, 10768. https://doi.org/10.1038/srep10768.

30. Wang, W., Li, C., Cho, M., and Ahn, S.-H. (2018). Soft Tendril-Inspired Grippers: Shape Morphing of Programmable Polymer–Paper Bilayer Composites. ACS Appl. Mater. Interfaces *10*, 10419–10427. https://doi.org/10.1021/acsami.7b18079.

31. Mitchell, S.K., Wang, X., Acome, E., Martin, T., Ly, K., Kellaris, N., Venkata, V.G., and Keplinger, C. (2019). An Easy-to-Implement Toolkit to Create Versatile and High-Performance HASEL Actuators for Untethered Soft Robots. Advanced Science *6*, 1900178. https://doi.org/10.1002/advs.201900178.

32. Must, I., Sinibaldi, E., and Mazzolai, B. (2019). A variable-stiffness tendril-like soft robot based on reversible osmotic actuation. Nat Commun *10*, 344. https://doi.org/10.1038/s41467-018-08173-y.

33. Zhang, J., Hu, Y., Li, Y., Ma, K., Wei, Y., Yang, J., Wu, Z., Rajabi, H., Peng, H., and Wu, J. (2022). Versatile Like a Seahorse Tail: A Bio-Inspired Programmable Continuum Robot For Conformal Grasping. Advanced Intelligent Systems, 2200263. https://doi.org/10.1002/aisy.202200263.

34. Webster, R.J., and Jones, B.A. (2010). Design and Kinematic Modeling of Constant Curvature Continuum Robots: A Review. The International Journal of Robotics Research *29*, 1661–1683. https://doi.org/10.1177/0278364910368147.

35. Todorov, E., Erez, T., and Tassa, Y. (2012). MuJoCo: A physics engine for model-based control. In 2012 IEEE/RSJ International Conference on Intelligent Robots and Systems, pp. 5026–5033. https://doi.org/10.1109/IROS.2012.6386109.

36. Packard, A., and Sanders, G.D. (1971). Body patterns of Octopus vulgaris and maturation of the response to disturbance. Animal Behaviour *19*, 780–790. https://doi.org/10.1016/S0003-3472(71)80181-1.

37. Hirose, S., and Umetani, Y. (1978). The development of soft gripper for the versatile robot hand. Mechanism and Machine Theory *13*, 351–359. https://doi.org/10.1016/0094-114X(78)90059-9.

38. Ruotolo, W., Brouwer, D., and Cutkosky, M.R. (2021). From grasping to manipulation with gecko-inspired adhesives on a multifinger gripper. Sci. Robot. *6*, eabi9773. https://doi.org/10.1126/scirobotics.abi9773.

39. Wang, Z., and Freris, N.M. (2024). Exploiting Frictional Effects to Reproduce Octopus-Like Reaching Movements with a Cable-Driven Spiral Robot. In 2024 IEEE 7th International Conference on Soft Robotics (RoboSoft) (IEEE), pp. 537–542. https://doi.org/10.1109/RoboSoft60065.2024.10522036.

40. Becker, K., Teeple, C., Charles, N., Jung, Y., Baum, D., Weaver, J.C., Mahadevan, L., and Wood, R. (2022). Active entanglement enables stochastic, topological grasping. Proc. Natl. Acad. Sci. U.S.A. *119*, e2209819119. https://doi.org/10.1073/pnas.2209819119.

41. Jones, T.J., Jambon-Puillet, E., Marthelot, J., and Brun, P.-T. (2021). Bubble casting soft robotics. Nature *599*, 229–233. https://doi.org/10.1038/s41586-021-04029-6.

42. George Thuruthel, T., Ansari, Y., Falotico, E., and Laschi, C. (2018). Control Strategies for Soft Robotic Manipulators: A Survey. Soft Robotics *5*, 149–163. https://doi.org/10.1089/soro.2017.0007.

43. Schulz, A.K., Boyle, M., Boyle, C., Sordilla, S., Rincon, C., Hooper, S., Aubuchon, C., Reidenberg, J.S., Higgins, C., and Hu, D.L. (2022). Skin wrinkles and folds enable asymmetric stretch in the elephant trunk. Proc. Natl. Acad. Sci. U.S.A. *119*, e2122563119. https://doi.org/10.1073/pnas.2122563119.